\makeatletter\def\graphicscache@inhibit{true}\makeatother

\documentclass[letterpaper, 10 pt, conference]{ieeeconf}  %

\IEEEoverridecommandlockouts                              %

\usepackage{cite}
\usepackage{amsmath,amssymb,amsfonts}
\usepackage{algorithmic}
\usepackage{graphicx}
\usepackage[dpi=500,qfactor=0.05]{graphicscache}
\usepackage{textcomp}
\usepackage{xcolor}

\usepackage{makeidx}

\usepackage{pgfplots}
\pgfplotsset{compat=newest}
\usepackage{multirow}
\usepackage{tabularx}
\usepackage{threeparttable}
\usepackage{booktabs}
\usepackage{array}
\usepackage{float}
\usepackage{graphbox}
\usepackage{tikz,tikz-3dplot}
\usetikzlibrary{fit,arrows,arrows.meta,automata,backgrounds,calc,chains,%
decorations.markings,decorations.pathreplacing,decorations.pathmorphing,%
matrix,positioning,shapes,shapes.geometric,shapes.symbols,spy,trees,tikzmark}
\usepackage{balance}
\usepackage[binary-units=true,product-units=single,per-mode=symbol,range-units=single,range-phrase=\,--\,,detect-all]{siunitx}
\DeclareSIUnit\pixel{px}
\usepackage{bm}
\usepackage{acronym}
\usepackage{tablefootnote}
\usepackage{hyperref}

\let\labelindent\relax
\usepackage[inline]{enumitem}

\definecolor{bg_color}{RGB}{95,95,95}

\let\vec\bm
\newcommand{\norm}[1]{\left\lVert#1\right\rVert}

\DeclareMathOperator*{\argmin}{arg\,min}

\newcommand{\reffig}[1]{Fig.~\ref{#1}}
\newcommand{\reftab}[1]{Tab.~\ref{#1}}
\newcommand{\refsec}[1]{Sec.~\ref{#1}}

\newcommand{\etal}{et al.~}

\newcommand{\cf}{cf.\ }

\usepackage{adjustbox}
\newcolumntype{R}[2]{%
    >{\adjustbox{angle=#1,lap=\width-(#2)}\bgroup}%
    l%
    <{\egroup}%
}
\newcolumntype{L}[1]{>{\raggedright\let\newline\\\arraybackslash\hspace{0pt}}m{#1}}

\title{\LARGE \bf
External Camera-based Mobile Robot Pose Estimation\\for Collaborative Perception with Smart Edge Sensors
}

\author{Simon Bultmann, Raphael Memmesheimer, and Sven Behnke%
\thanks{This work was funded by grant BE 2556/16-2 (Research Unit FOR 2535 Anticipating Human Behavior) of the German Research Foundation (DFG)}%
\thanks{All authors are with the Autonomous Intelligent Systems group, %
		University of Bonn, Germany;
        {\tt\small bultmann@ais.uni-bonn.de}}%
}

\begin{document}

\maketitle
\thispagestyle{empty}
\pagestyle{empty}

\begin{tikzpicture}[remember picture,overlay]
\node[anchor=north west,align=left,font=\sffamily,yshift=-0.2cm,xshift=0.2cm] at (current page.north west) {%
  Accepted for IEEE International Conference on Robotics and Automation (ICRA), London, UK, June 2023.
};
\end{tikzpicture}%
\vspace{-0.2cm}%

\begin{abstract}
We present an approach for estimating a mobile robot's pose w.r.t. the allocentric coordinates of a network of static cameras using multi-view RGB images. The images are processed online, locally on smart edge sensors by deep neural networks to detect the robot and estimate 2D keypoints defined at distinctive positions of the 3D robot model. Robot keypoint detections are synchronized and fused on a central backend, where the robot's pose is estimated via multi-view minimization of reprojection errors.
Through the pose estimation from external cameras, the robot's localization can be initialized in an allocentric map from a completely unknown state (\textit{kidnapped robot problem}) and robustly tracked over time.
We conduct a series of experiments evaluating the accuracy and robustness of the camera-based pose estimation compared to the robot's internal navigation stack, showing that our camera-based method achieves pose errors below \SI{3}{\centi\meter} and \SI{1}{\degree} and does not drift over time, as the robot is localized allocentrically.
With the robot's pose precisely estimated, its observations can be fused into the allocentric scene model. We show a real-world application, where observations from mobile robot and static smart edge sensors are fused to collaboratively build a 3D semantic map of a $\sim$\SI{240}{\square\meter} indoor environment.
\end{abstract}

\section{Introduction}
\label{sec:Introduction}
Semantic scene understanding is an important requirement for intelligent robot action, such as collision-free navigation, object manipulation, or human-robot interaction. Scene interpretation from a single sensor view, however, has a limited field of perception. %
Collaborative perception between mobile robots and distributed static smart edge sensors alleviates this issue and enables to build 3D semantic models of large scenes without being limited by the measurement range or occlusions of a single sensor.

A key prerequisite for fusing observations from different perspectives is knowing the relative sensor poses.
While the extrinsic calibration of static sensors can be performed beforehand, a mobile sensor's pose w.r.t. %
the sensor network must be initialized and tracked to fuse the robot observations into the allocentric model in a consistent manner.

In this work, we propose to estimate a mobile robot's pose w.r.t. the allocentric coordinates of a network of static smart cameras using multi-view RGB images. For this, we build upon our previous work on 3D semantic scene perception using distributed smart edge sensors~\cite{bultmann_ias2022}, where images are processed locally on the sensor boards by deep neural networks for semantic image interpretation.
For robot pose estimation, we process the images with Convolutional Neural Networks (CNNs) for robot detection and estimation of 2D projections of keypoints defined at distinctive positions of the 3D robot model. Unlike classical robot-to-camera pose estimation systems, our method does not require attaching fiducial markers to the robot. Furthermore, the CNN for keypoint estimation is trained only on synthetic data obtained through randomized scene generation~\cite{stilleben_2020}. %
Robot keypoint detections are streamed to a central backend where they are synchronized and the robot's pose is estimated via multi-view minimization of reprojection errors.
Through the pose estimation from external camera views, the robot's localization is initialized in the global map from a completely unknown state, and robustly tracked over time.
Furthermore, using multiple sources for localization, i.e. the external camera views together with the robot's internal 2D LiDAR-based navigation, increases robustness in highly cluttered, dynamic real-world environments, where few distinct features, such as walls or columns, are visible in the LiDAR due to occlusions.

With the robot's pose precisely estimated, it is integrated as a mobile sensor node into the camera network. We deploy the CNNs for semantic image interpretation from~\cite{bultmann_ias2022} onto the robot's inference accelerator and use its RGB-D camera to obtain semantically annotated point clouds of the robot's view. The robot's observations are then fused into the allocentric scene model to build a 3D semantic map of a large room in collaboration with the static smart edge sensors.

\begin{figure}[t]
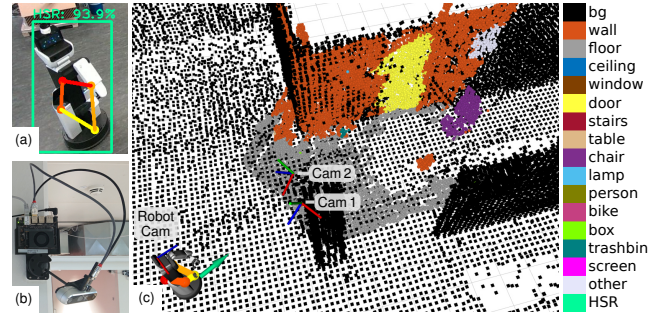

	\centering
		\begin{tikzpicture}
		    \node[inner sep=0,anchor=north west] (image1) at (0, 0){\includegraphics[height=2.cm,trim= 18 0 363 120, clip]{figures/teaser_44s_d455_3_overlay.png}};
		    
		    \node[inner sep=0,anchor=north west,yshift=-0.1cm] (image2) at (image1.south west) {\includegraphics[height=2.cm,trim= 3000 900 250 550, clip]{figures/DSC_3168.JPG}};
		    
			\node[inner sep=0,anchor=north west,xshift=0.05cm] (image3) at (image1.north east) {\includegraphics[height=4.1cm,trim=20 0 0 90, clip]{figures/teaser_44s_3D_4.png}};
			
			\node[inner sep=0,anchor=north west,xshift=0.05cm] (image4) at (image3.north east) {\includegraphics[height=4.1cm]{figures/legend_ade20k_indoor_reduced_hsr.pdf}};

			\node[label,scale=.75, anchor=south west,xshift=-0.07cm, rectangle, fill=white, align=center, font=\scriptsize\sffamily] (n_0) at (image1.south west) {(a)};
			\node[label,scale=.75, anchor=south west,xshift=-0.07cm, rectangle, fill=white, align=center, font=\scriptsize\sffamily] (n_1) at (image2.south west) {(b)};
			\node[label,scale=.75, anchor=south west, rectangle, fill=white, align=center, font=\scriptsize\sffamily] (n_3) at (image3.south west) {(c)};
			\node[label, scale=.75, anchor=south west, xshift=5.25cm, yshift=-0.37cm, rectangle, rounded corners=2, inner sep=0.06cm, fill={white!90!black},opacity=.9,text opacity=1, align=center, font=\scriptsize\sffamily] (l_cam1) at (image2.north west) {Cam\,2};
			\draw[{white!90!black},opacity=.9, very thick] (l_cam1.180) ++(0.006, 0) -- ++(-0.15, 0);
    \node[label, scale=.75, anchor=south west, xshift=5.35cm, yshift=-0.88cm, rectangle, rounded corners=2, inner sep=0.06cm, fill={white!90!black},opacity=.9,text opacity=1, align=center, font=\scriptsize\sffamily] (l_cam2) at (image2.north west) {Cam\,1};
    \draw[{white!90!black},opacity=.9, very thick] (l_cam2.180) ++(0.006, 0) -- ++(-0.12, 0);
    
    \node[label, scale=.75, anchor=south west, xshift=2.15cm, yshift=-1.4cm, rectangle, rounded corners=2, inner sep=0.06cm, fill={white!90!black},opacity=.9,text opacity=1, align=center, font=\scriptsize\sffamily] (l_cam2) at (image2.north west) {Robot\\Cam};
    \draw[{white!90!black},opacity=.9, very thick] (l_cam2.270) ++(0, 0.006) -- ++(0., -0.2);
		\end{tikzpicture}
		\vspace{-.7em}
	\caption{Robot pose estimation for collaborative perception: (a) Toyota Human Support Robot (HSR) employed in this work, with 2D bounding box and keypoints used for pose estimation detected by static smart edge sensors (b). (c) 3D scene view with robot model and 3D keypoints at the estimated robot pose (green arrow) and robot's semantic observations (colored point cloud) to be fused into the allocentric scene model (black squares). Robot observations fit the allocentric model, showing that the robot's pose is initialized globally coherent through localization by the smart edge sensors.}
	\label{fig:teaser}
	\vspace{-1em}
\end{figure}

Fig.~\ref{fig:teaser} shows the employed Toyota Human Support Robot (HSR) and static smart edge sensors together with a 3D view of the robot's semantic observations to be fused into the 3D scene model, initialized from a prior without any semantic annotations. As the robot pose is initialized globally coherently through the external keypoint-based pose estimation, its observations consistently fit the allocentric model.\\
In summary, the main contributions of this paper are:
\begin{itemize}
  \item A novel method for marker-less mobile robot pose estimation using multi-view keypoint detections to initialize the robot's localization in a global map (\textit{kidnapped robot problem}) and robustly track it over time,
  \item integration of a mobile robot into a network of static smart edge sensors, fusing the robot's observations from changing viewpoints into the global scene model,
  \item quantitative evaluation of the pose estimation accuracy and robustness, and
  \item demonstration of collaborative perception in real-world scenes between mobile robot and sensor network to build a globally consistent 3D semantic map.
\end{itemize}

\section{Related Work}
\label{sec:Related_Work}
Visual robot detection and localization have been research topics of high interest for 
a variety of autonomous systems like drones~\cite{ashraf2021dogfight},
mobile robot platforms~\cite{pizarro2008mobile,pizarro2010localization, meger2009inferring},
humanoids~\cite{amini2022real}, autonomous cars~\cite{gawel2018x}, and underwater
robots~\cite{shkurti2017underwater, joshi2020deepurl}. External localization of moving objects is 
also employed for traffic monitoring~\cite{strand2022systematic, severi2018beyond}
or action recognition~\cite{wang2016collaborative}.

Detection methods localize robots in images and allow for tracking applications like underwater convoying~\cite{shkurti2017underwater} or monitoring~\cite{ashraf2021dogfight}.
In contrast, we aim to estimate a robot's pose from multiple views to initialize its localization in a map and to compensate for localization errors in highly cluttered, dynamic environments.
Gawel~\etal\cite{gawel2018x} presented a semantic graph-based multi-view approach 
for robot localization in autonomous driving scenarios. They use semantic 
estimates from various views like front and aerial views and integrate them in separate
target and query graphs. %
Lu~\etal\cite{lu2022pose} presented an approach for marker-less robot pose estimation via 
keypoint detection~\cite{mathis2018deeplabcut}.
They aim to define keypoints on the robot maximizing the localization performance in 2D and 3D.
Their approach utilizes synthetic training data and transfers well to real sensor data.
They estimate the pose of stationary robots from single views, whereas our approach globally tracks
a mobile manipulation platform.
Lee~\etal\cite{lee2020camera} consider the inverse problem of localizing the camera in
relation to an articulated robot. They estimate {2D} keypoints of robot joints 
and finally recover the camera extrinsics w.r.t. the robot manipulator.
Pizzaro~\etal\cite{pizarro2008mobile, pizarro2010localization} presented 
approaches for robot localization from a single view.
Shim~\etal\cite{shim2015mobile} proposed an approach for multi-view mobile robot localization. 
They first project the camera images onto a common plane and then localize the robot using its contours after removing shadows. 
Similarly, Chakravarty and Jarvis~\cite{chakravarty2009external} utilize a surveillance camera system
to localize the robot in the image plane by background subtraction of a color thresholded image. 
Similar to our work, the surveillance camera view is extended by the robot's view.
Obviously, such approaches would fail in highly dynamic and cluttered environments.

The field-of-view of autonomous mobile robots can be extended in comparison to single-view systems
using smart edge sensors for global collaborative perception.
In prior work, we utilized smart edge sensors to estimate 3D human poses~\cite{Bultmann_RSS_2021} from multiple views and integrate 3D semantic perception~\cite{bultmann_ias2022}.
In this work, we focus on the integration of a mobile robot as an additional smart edge sensor node, laying the foundations for many potential applications like %
external, camera-based control and task planning as well as augmentation of 
the internal robot view by the integration of human poses or semantic segmentation estimates
that are robust against occlusions. %
Rekleitis~\etal\cite{rekleitis2001collaborative} presented an approach for 
collaborative exploration of visual maps using two mobile robots of different 
capability levels, where one robot was actively collecting visual data for mapping and the second, passive robot visually refined the pose of the moving robot.
Dong~\etal\cite{dong2019multi} proposed a multi-robot collaborative approach for
dense reconstruction. They associate task views of uncertain or unexplored map areas to the robots based on the traveling salesman problem.
In contrast to our approach, they concentrate on an actively moving set of mobile robot platforms
and omit semantics.
Similarly, Yue~\etal\cite{yue2020collaborative}, propose to combine an Unmanned Aerial Vehicle (UAV) and an Unmanned Ground Vehicle (UGV) for collaborative semantic mapping.
Ahmed~\etal\cite{ahmed2021towards} propose to integrate top-view surveillance 
cameras for collaborative robotics but assume that the estimations from surveillance 
cameras are provided to the robot without actively fusing information of the robot's
sensors.

With PoseCNN~\cite{posecnn}, a 6D pose estimation approach jointly estimating semantic labels on a pixel level has been presented. Xiang~\etal estimate the translation by regressing the object center in camera coordinates and separately regress the rotation using a newly introduced ShapeMatch-Loss.
Wang~\etal\cite{wang2019densefusion} presented a dense fusion method for 6D pose estimation. They first estimate semantic segmentation and bounding boxes, and then densely fuse RGB and depth images in an embedding. Finally, an iterative pose integrator refines the 6D pose.
Keypoint-based approaches for 6D pose estimation adopt a two-stage pipeline, first estimating keypoints for each object instance and, second, retrieving the object poses via PnP~\cite{EPnP}. They have been widely used in recent years~\cite{BB8,PVNet,zappel20226d,yolopose_2022,lee2020camera}.
Although most pose estimation approaches are object-centric, in this paper, we propose to use a keypoint-based estimator for the visual pose estimation of a mobile robot.
 
\section{Method}
\label{sec:Method}
We consider scenarios where $N$ externally mounted cameras $C_i, i = 1,\ldots,N$ observe a mobile robot from different viewpoints.
The intrinsic and extrinsic calibration of the external cameras are performed beforehand~\cite{paetzold_camcalib_2022}, i.e., we assume the transformation $^{C_i}_W\vec{T} \in \mathbb{R}^{4\times4}$ from world to camera coordinates and the projection $\Pi_i(\cdot)$ to the image plane of $C_i$ to be known.
The cameras observe 2D projections $\vec{k}_{ij} \in \mathbb{R}^2$ of $M$ keypoints $\vec{p}_j \in \mathbb{R}^3, j = 1, \ldots, M$ defined on distinct positions of the 3D robot model.
From these observations, we aim to estimate the robot's pose in world coordinates $^{W}_R\vec{T} \in \mathbb{R}^{4\times4}$. As the employed mobile robot moves on the ground plane, we consider three degrees of freedom (DoF): $\vec{x} = [x, y]^\top \in\mathbb{R}^2$, and $\theta\in (-\pi, \pi]$, with
\begin{align}
^{W}_R\vec{T}\left(\vec{x}, \theta\right) = \begin{bmatrix} \vec{R}\left(\theta\right) & \vec{0} & \vec{x} \\ \vec{0}^\top & 1 & 0 \\ \vec{0}^\top & 0 & 1 \end{bmatrix}
, \vec{R}\left(\theta\right) \in \mathcal{SO}\left(2\right)\,.
\label{eqn:robot_pose}
\end{align}
Extending this formulation to more DoFs is straightforward. %
We estimate the robot pose in a two-stage process, similar to Lee~\etal\cite{lee2020camera} but exploiting multi-view observations: First, robot keypoints are detected on the images, locally on the sensor boards (\refsec{sec:kps}), and, second, the robot pose is estimated from a synchronized set of observations via multi-view minimization of reprojection errors (\refsec{sec:pose}).

\subsection{Robot Keypoint Detection}
\label{sec:kps}
We aim to estimate the robot's base pose, independent of the articulation of the robot head or arm, and therefore define the keypoints used for pose estimation on distinct points of the rigid robot base (see \reffig{fig:train_data} (c)).
Keypoints defined manually at distinct points of the 3D model were shown to perform better than automatically defined ones~\cite{zappel20226d}.

\subsubsection{Network Architecture}
We employ a top-down approach for robot keypoint detection, i.e., first detecting a bounding box of the robot in the full image ($848\times480$\,px) and then estimating keypoints on the crop of the robot. Top-down methods achieve better scale invariance than estimating keypoints on the full input image, as the crop of a detected robot is interpolated to a fixed resolution for keypoint estimation (here $192\times256$\,px).
Camera images are processed locally on the smart edge sensors, using a Nvidia Jetson Xavier NX embedded inference accelerator (\cf\reffig{fig:teaser} (b)). We employ the CNN architectures used in prior work for efficient person keypoint estimation on the embedded hardware~\cite{Bultmann_RSS_2021,bultmann_ias2022}: The recent MobileDet architecture~\cite{xiong_mobiledets_2021} is used for robot detection and the network of Xiao~\etal\cite{xiao_simple_2018} with a MobileNet\,V3~\cite{mobilenetv32019} backbone for keypoint estimation.

\subsubsection{Training Data}
To reduce the labeling effort to a minimum, we train the networks predominantly on synthetic data.
The CNN for keypoint estimation is trained purely on simulated data (36k samples), while we combine synthetic data and manually annotated real images (12k resp. 3.5k samples) for robot detection.
The combination of real and synthetic data helps to boost detector performance in highly cluttered real-world environments and bounding-box labels are less costly to obtain than keypoint annotations. The keypoint CNN generalizes well from only synthetic data.

\begin{figure}
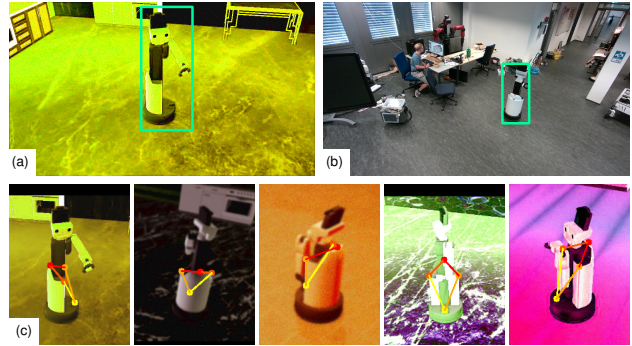

\centering
\begin{tikzpicture}
		    \node[inner sep=0,anchor=north west] (image1) at (0, 0){\includegraphics[height=2.3cm,trim= 0 0 0 0, clip]{figures/train_det_synth_696045.png}};
		    
		    \node[inner sep=0,anchor=north west,xshift=0.1cm] (image2) at (image1.north east) {\includegraphics[height=2.3cm,trim= 0 0 0 0, clip]{figures/train_det_real_106764.png}};
		    
			\node[inner sep=0,anchor=north west,yshift=-0.1cm] (image3) at (image1.south west) {\includegraphics[height=2.13cm,trim=0 0 0 0, clip]{figures/train_synth_696046_crop.png}};
			\node[inner sep=0,anchor=north west,xshift=0.05cm] (image4) at (image3.north east) {\includegraphics[height=2.13cm,trim=0 0 0 0, clip]{figures/train_synth_687061_crop.png}};
			 \node[inner sep=0,anchor=north west,xshift=0.05cm] (image5) at (image4.north east) {\includegraphics[height=2.13cm,trim=0 0 0 0, clip]{figures/train_synth_685104_crop.png}};
			 \node[inner sep=0,anchor=north west,xshift=0.05cm] (image6) at (image5.north east) {\includegraphics[height=2.13cm,trim=0 0 0 0, clip]{figures/train_synth_699042_crop.png}};
			 \node[inner sep=0,anchor=north west,xshift=0.05cm] (image7) at (image6.north east) {\includegraphics[height=2.13cm,trim=0 0 0 0, clip]{figures/train_synth_crop_691077_crop.png}};
			
			\node[label,scale=.75, anchor=south west,xshift=-0.07cm, rectangle, fill=white, align=center, font=\scriptsize\sffamily] (n_0) at (image1.south west) {(a)};
			\node[label,scale=.75, anchor=south west,xshift=-0.07cm, rectangle, fill=white, align=center, font=\scriptsize\sffamily] (n_1) at (image2.south west) {(b)};
			\node[label,scale=.75, anchor=south west, rectangle, fill=white, align=center, font=\scriptsize\sffamily] (n_3) at (image3.south west) {(c)};
		\end{tikzpicture}
\vspace{-.7em}
\caption{Examples of training images: (a) synthetic and (b) real images used for the detector; (c) synthetic images used for the keypoint estimation.}
\vspace{-1em}
\label{fig:train_data}
\end{figure}

We employ an extension of the \textit{stilleben}-framework~\cite{stilleben_2020,sl-cutscenes} for photorealistic, randomized scene rendering to generate multiple scenes of our robot moving through varying indoor environments, using the 3D robot model.
Randomizing scene parameters in addition to image augmentation helps to bridge the reality gap for sim-to-real transfer. %
\reffig{fig:train_data} shows examples of the employed training data.
Note, that the network also learns to detect occluded keypoints from the surrounding image context.

\subsection{Robot Pose Estimation}
\label{sec:pose}
2D robot keypoints are sent over a network to a central backend, where detections from multiple cameras are software-synchronized via their timestamps and associated to corresponding frame-sets. %
The robot pose ${^{W}_R\vec{T}}$ \eqref{eqn:robot_pose} is then recovered by solving a weighted nonlinear least squares problem via minimization of multi-view reprojection errors:
\begin{align}
{^{W}_R\vec{T}} = \argmin_{^{W}_R\vec{T}} \sum_{i=1}^N \sum_{j=1}^M w_{ij} \norm{\vec{k}_{ij} - \Pi_i\left({^{C_i}_W\vec{T}} {^{W}_R\vec{T}} \vec{p}_j\right)}^2\mkern-6mu,
\label{eqn:opt}
\end{align}
with weights $w_{ij}$ depending on the confidence of the keypoint detection in the respective camera, similar to~\cite{Bultmann_RSS_2021}.
We use the Levenberg-Marquardt algorithm as implemented in the Ceres library~\cite{Agarwal_Ceres_Solver_2022} for optimization.
We discern two different ways to initialize the optimization:
\begin{enumerate*}[label=~(\roman*),before=\\,itemjoin=\\]
\item When the robot pose has been initialized in the global map and a current pose estimate is available from its internal navigation stack, we use this to initialize the optimization.
\item When no prior estimate is available, e.g., to initialize the robot's pose or when the communication link to the robot is unavailable, we use the PnP algorithm~\cite{EPnP} to obtain pose candidates from each individual camera view with $M\geq 4$ detected keypoints. The candidate poses are transformed to world coordinates using $^{C_i}_W\vec{T}^{-1}$ and projected to the ground plane.
The initialization for Eq.~\eqref{eqn:opt} is then obtained via interpolation between the candidates, using spherical linear interpolation for the orientation.
\end{enumerate*}

In our experiments (\refsec{sec:Evaluation}), we observe that the pose estimation is very well constrained when the robot is detected in $N\geq2$ cameras, but less stable when visible in only a single camera and far away from the camera. This is due to the robot body having a small width of \SI{35}{\centi\meter}, providing only a narrow baseline for orientation and depth estimation.
To alleviate this issue, we implement a simple yet effective bearing-only heuristic for outlier detection:
For pose estimates from a single camera view, we prevent unrealistically high changes in orientation or distance to the camera above a threshold $d_\text{\ensuremath{\theta}}$ resp. $d_\text{depth}$ by updating only the translation orthogonal to the camera's viewing direction.

Lastly, we employ a pose graph~\cite{Dellaert} to fuse the absolute pose estimations from external cameras, which occur sparsely, at waypoints where the robot is static (\cf\refsec{sec:collab}), with the continuous robot-internal odometry.
The pose estimates from the robot's internal navigation stack are inserted into the pose graph as binary odometry constraints, while the pose estimates from external cameras are used as absolute, unary constraints. The covariance of the camera pose estimates is proportional to the reprojection error residual~\eqref{eqn:opt} and inversely to the number of cameras used in the optimization, giving the highest confidence to consistent estimates with low reprojection error in a large number of cameras. The pose graph is optimized each time a new external camera pose estimate occurs.

\subsection{Collaborative Localization and Perception}
\label{sec:collab}
\begin{figure}[t]
  \centering
  \resizebox{1.0\linewidth}{!}{%
\begin{tikzpicture} 
[content_node/.append style={font=\sffamily,minimum size=1.5em,minimum width=6em,draw,align=center,rounded corners,scale=0.65},
label_node/.append style={font=\sffamily,scale=0.6},
group_node/.append style={font=\sffamily,dotted,align=center,rounded corners,inner sep=1em,thick},>={Stealth[inset=0pt,length=4pt,angle'=45]}]
\tikzset{junction/.append style={circle, fill=black, minimum size=3pt, draw}}

\definecolor{red}{rgb}     {0.9,0.0,0.0}
\definecolor{green}{rgb}   {0.0,0.5,0.0}
\definecolor{blue}{rgb}    {0.0,0.0,0.5}
\definecolor{grey}{rgb}    {0.5,0.5,0.5}

\node(SmartEdge_Group_Label)[label_node,anchor=south west] at (-2.3em, 1.8em) {\textbf{Smart Edge Sensor Jetson NX $1,\,\ldots,\,N$}};
\draw[thick, rounded corners, grey!20!white,fill] (-2.3em,1.8em) -- (23.em,1.8em) -- (23.em,-1.3em) -- (-2.3em,-1.3em) -- cycle;
\node(Camera)[content_node,fill=green!15!white] at (0, 0) {RGB-D\\Camera};
\node(Robot_kps)[content_node,fill=blue!15!white, anchor=north west, xshift = 3.em] at (Camera.north east) {Robot Keypoint\\Detection};
\node(RGB_Segm_Detection)[content_node,fill=blue!15!white, anchor=north west, xshift = 3.em] at (Robot_kps.north east) {RGB Detection and\\ Segmentation~\cite{bultmann_ias2022}};
\node(Pcd_Fusion)[content_node,fill=blue!15!white, anchor=north west, xshift = 3.em] at (RGB_Segm_Detection.north east) {Point Cloud\\Fusion~\cite{bultmann_ias2022}};

\draw[dotted, very thick] (Pcd_Fusion) ++(3.2em, 0.em) -- ++(.9em, 0.em);

\draw[thick, rounded corners, grey!20!white,fill] (2.em,-3.em) -- (23.5em,-3.em) -- (23.5em,-8.em) -- (2.em,-8.em) -- cycle;
\node(Backend_Group_Label)[label_node,anchor=south west] at (2.em,-3.7em) {\textbf{Backend}};
\node(Pose_Est)[content_node,fill=blue!15!white, minimum height=3.5em, anchor=north west, yshift = -5em, xshift = 1em] at (Robot_kps.south west) {Robot Pose\\Estimation};
\node(Pose_Graph)[content_node,fill=blue!15!white, anchor=north west, yshift = -7em] at (RGB_Segm_Detection.south west) {Pose Graph\\Fusion};
\node(Prior_Map_3D)[content_node,fill=blue!15!white, anchor=north west, xshift = 4em, yshift = -3.3em] at (RGB_Segm_Detection.south west) {3D Prior Map};
\node(Sem_Map)[content_node,fill=blue!15!white, minimum height=5em, anchor=north west, yshift = -4.5em] at (Pcd_Fusion.south west) {3D Semantic\\Mapping};

\node(HSR_Group_Label)[label_node,anchor=south west] at (-1.5em,-9.3em) {\textbf{HSR Robot}};
\draw[thick, rounded corners, grey!20!white,fill] (-1.5em,-9.3em) -- (23.5em,-9.3em) -- (23.5em,-16.3em) -- (-1.5em,-16.3em) -- cycle;

\node(HSR_Main_Group_Label)[label_node,anchor=south west] at (-1.2em,-10.2em) {\textbf{Main PC}};
\node(HSR_Jetson_Group_Label)[label_node,anchor=south west] at (10.3em,-10.2em) {\textbf{Jetson TX2 accelerator}};
\draw[thick, rounded corners, grey!60!white] (-1.2em,-10.2em) -- (9.em,-10.2em) -- (9.em,-16.em) -- (-1.2em,-16.em) -- cycle;
\draw[thick, rounded corners, grey!60!white] (10.3em,-10.2em) -- (23.2em,-10.2em) -- (23.2em,-16.em) -- (10.3em,-16.em) -- cycle;

\node(RGB_Segm_Detection_HSR)[content_node,fill=blue!15!white, anchor=north west, yshift = -15.em] at (RGB_Segm_Detection.south west) {RGB Detection and\\ Segmentation~\cite{bultmann_ias2022}};
\node(Pcd_Fusion_HSR)[content_node,fill=blue!15!white, anchor=north west, yshift = -15.em] at (Pcd_Fusion.south west) {Point Cloud\\Fusion~\cite{bultmann_ias2022}};
\node(Camera_HSR)[content_node,fill=green!15!white, anchor=north west, xshift = 1.5em, yshift = -2.em] at (RGB_Segm_Detection_HSR.south west) {Head RGB-D\\Camera};

\node(Robot_Nav_Stack)[content_node,fill=blue!15!white, minimum height=5.em, anchor=north west, yshift = -15.em] at (Robot_kps.south west) {Navigation\\Stack};

\node(LiDAR)[content_node,fill=green!15!white, anchor=north west, xshift = -7.5em, yshift = -.5em] at (Robot_Nav_Stack.north west) {2D LiDAR};
\node(IMU)[content_node,fill=green!15!white, anchor=north west, xshift = -7.5em, yshift = -3.em] at (Robot_Nav_Stack.north west) {IMU};
\node(Prior_Map_2D)[content_node,fill=blue!15!white, anchor=north west, yshift = -1.em] at (Robot_Nav_Stack.south west) {2D Prior Map};

\draw[->, thick] (Camera.351) -- node[label_node,near start,below] {\;\;RGB} (Camera.351 -| Robot_kps.180);
\draw[->, thick] (3.2em, 0 |- Camera.351) |- ++(6.5em, 1.5em) |- (RGB_Segm_Detection.180);
\node(junct_0)[junction, anchor=center, scale=0.3] at (3.2em, 0 |- Camera.351){};

\draw[->, thick] (Camera.9) -- ++(.6em, 0) |- node[label_node,midway,left] {Depth} ++(15.em, 1.15em) |- (Pcd_Fusion.167);
\draw[->, thick] (RGB_Segm_Detection.353) -- (RGB_Segm_Detection.353 -| Pcd_Fusion.180);

\draw[->, thick] (Robot_kps.270) -- node[label_node,near start,left,text width=5em,align=right] {2D Robot\\Keypoints} (Robot_kps.270 |- Pose_Est.90);
\draw[->, dashed, thick] (Robot_kps.0) + (0,-2em) -| (Pose_Est.60);
\draw[->, thick] (Pcd_Fusion.270) -- node[label_node,near start,left,text width=7em,align=right] {Semantic\\Point Cloud}(Pcd_Fusion.270 |- Sem_Map.90);
\draw[->, dashed, thick] (Pcd_Fusion.0) + (0,-2em) -| (Sem_Map.60);

\draw[->, thick] (Prior_Map_3D.7) -| (Sem_Map.120);

\draw[->, thick, red] (Pose_Est.270 -| Robot_Nav_Stack.90) -- node[label_node,near start,left,text width=10em,align=right] {Pose Initialization;\\Pose Correction\\Feedback}(Robot_Nav_Stack.90);
\draw[->, thick, red] (Robot_Nav_Stack.90 |- Pose_Graph.190) -- ++(3em, 0) |- (Pose_Graph.170);
\node(junct_0)[junction, red, anchor=center, scale=0.3] at (Robot_Nav_Stack.90 |- Pose_Graph.190){};

\draw[->, thick, red] (Pose_Graph.0) -- node[label_node,midway,above] {Fused Robot} node[label_node,midway,below] {Pose}(Pose_Graph.0 -| Sem_Map.180);

\draw[->, thick] (Robot_Nav_Stack.55) -- ++(0, 3em) -- node[label_node,near end,below] {\qquad Robot Odometry} ++(2.5em, 0) |- (Pose_Graph.190);

\draw[->, thick] (LiDAR.0) -- (LiDAR.0 -| Robot_Nav_Stack.180);
\draw[->, thick] (IMU.0) -- (IMU.0 -| Robot_Nav_Stack.180);
\draw[->, thick] (Prior_Map_2D.90) -- (Prior_Map_2D.90 |- Robot_Nav_Stack.270);

\draw[->, thick] (Camera_HSR.120) -- node[label_node,midway,left] {RGB} (Camera_HSR.120 |- RGB_Segm_Detection_HSR.270);
\draw[->, thick] (RGB_Segm_Detection_HSR.7) -- (RGB_Segm_Detection_HSR.7 -| Pcd_Fusion_HSR.180);
\draw[->, thick] (Camera_HSR.60) -- ++(0, .7em) -- node[label_node,midway,below] {Depth} ++(3.2em, 0) |- (Pcd_Fusion_HSR.190);

\draw[->, thick, red] (Pcd_Fusion_HSR.90) -- node[label_node,midway,left,text width=7em,align=right] {Robot Semantic\\Point Cloud}(Pcd_Fusion_HSR.90 |- Sem_Map.270);

\draw[->, thick] (Sem_Map.0) -- node[label_node,near start,above] {\qquad\quad\;3D Semantic} node[label_node,near start,below] {\qquad\quad\;Scene Model} ++(3.4em, 0);

\end{tikzpicture}
}
  \vspace{-2em}
  \caption{Overview of the proposed sensor network architecture for col\-la\-bo\-ra\-tive localization and perception. $N$ external smart edge sensors observe mobile HSR robot and scene from static viewpoints. Robot pose is initialized and corrected via external camera pose estimation. Robot observations from changing viewpoints are fused into the allocentric scene model.}
  \vspace{-1em}
  \label{fig:system}
\end{figure}
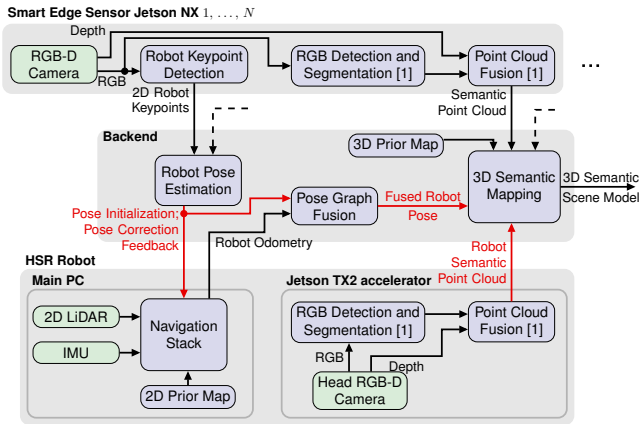
\reffig{fig:system} gives an overview of the proposed sensor network architecture for collaborative localization and perception.
The mobile HSR robot is integrated into a network of static smart edge sensors for 3D semantic scene perception~\cite{bultmann_ias2022}.
The robot pose is initialized in the allocentric scene from the external cameras.
During operation, external robot pose estimations are sent as pose-correction feedback to the robot sparsely at waypoints where the robot is static, updating the robot's internal particle filter-based localization~\cite{thrun2001robust} to keep it globally coherent within the scene model.
The interface provided by the robot's navigation stack permits to update the localization from external measurements only when not in movement.
To simplify system integration and for better interoperability, we use the robot's navigation stack as provided by the manufacturer and consider it as a closed module.
We do not aim to fully control the robot localization via the external cameras for the robot to keep its local autonomy: Its navigation stack integrates information from the camera network when available, but does not depend on it. When, e.g., the communication link is lost, the robot can still navigate autonomously using its integrated sensors.

For collaborative perception, the mobile robot provides changing viewpoints of areas outside the field-of-view of the static smart edge sensors, and its semantic percepts are fused into the allocentric scene model.

\subsubsection{Semantic Perception onboard the Robot}
The HSR's computing system comprises a main PC used for communication, robot navigation, and control and an embedded deep learning inference accelerator (Nvidia Jetson TX2). We employ the latter to deploy CNNs for object detection and semantic segmentation from~\cite{bultmann_ias2022} and obtain semantic point clouds using the Asus Xtion RGB-D camera mounted on the robot's head.
The robot's inference accelerator hardware is a previous generation of the compute boards of the static smart edge sensors, with lower, but sufficient resources for semantic point cloud estimation at \SI{1}{\hertz}.

\subsubsection{Semantic Map Fusion}
The robot's semantic observations from changing viewpoints are integrated into the allocentric 3D semantic map using the probabilistic fusion proposed in~\cite{bultmann_ias2022}. Because we consider the mobile robot's pose less reliable than the a-priori calibrated static sensor poses, we perform an ICP alignment step before integrating the robot's data into the map. This compensates for the drift accumulated between pose corrections and tolerances in the robot's forward kinematics. As a good initialization is critical for ICP, precise robot localization is necessary for initializing the alignment.
 
\section{Evaluation}
\label{sec:Evaluation}
\begin{figure}[t]
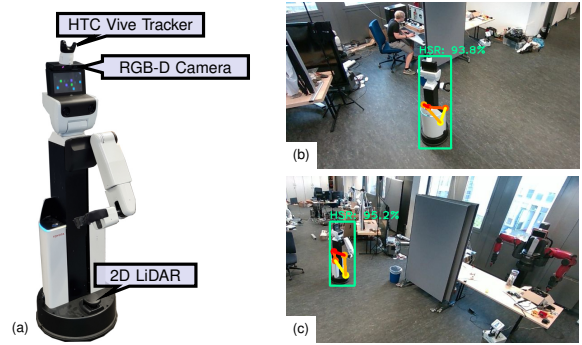

  \centering

  \begin{tikzpicture}
      [boxstyle/.style={font=\sffamily,black,fill=blue!20!white,fill opacity=0.8,text opacity=1,text=black,draw,very thick,align=center,rectangle callout}]
          \definecolor{red}{rgb}{0.7,0.0,0.0}
        \definecolor{blue}{rgb}{0.2,0.2,0.7}
        \node[anchor=north west, inner sep=0] (hsr) at (0, 0) {\includegraphics[trim=0px 0px 0px 0px,clip,height=4.cm]{figures/hsr.png}};

        \begin{scope}[shift=(hsr.south west),x={(hsr.south east)},y={(hsr.north west)}]
            \node[boxstyle, scale=0.6,text width=3.cm,callout relative pointer={(-0.4, -0.05)}] at (.95,1.05) {HTC Vive Tracker};
            \node[boxstyle, scale=0.6,text width=3.cm,callout relative pointer={(-0.35, 0)}] at (1.4,0.91) {RGB-D Camera};
            \node[boxstyle, scale=0.6,text width=2.0cm,callout relative pointer={(-0.3, -0.05)}] at (1.1,0.22) {2D LiDAR};
        \end{scope}
        \node[inner sep=0,anchor=north west,xshift=2.cm, yshift = .5cm] (image2) at (hsr.north east) {\includegraphics[trim=0px 0px 100px 57px,clip,height=2.2cm]{figures/hsr_det_d455_2.png}};
        \node[inner sep=0,anchor=north west,yshift=-.1cm] (image3) at (image2.south west) {\includegraphics[trim=120px 68px 0px 0px,clip,height=2.2cm]{figures/hsr_det_d455_3.png}};
        
        \node[label,scale=.75, anchor=south west, xshift =-.55cm, rectangle, fill=white, align=center, font=\scriptsize\sffamily] (n_0) at (hsr.south west) {(a)};
			\node[label,scale=.75, anchor=south west, rectangle, fill=white, align=center, font=\scriptsize\sffamily] (n_1) at (image2.south west) {(b)};
			\node[label,scale=.75, anchor=south west, rectangle, fill=white, align=center, font=\scriptsize\sffamily] (n_3) at (image3.south west) {(c)};
  \end{tikzpicture}
  \vspace{-.7em}
  \caption{Robot setup (a) and exemplary robot keypoint detections (b, c).}
  \vspace{-1.5em}
  \label{fig:experimental_setup}
\end{figure}

\begin{figure*}[!ht]
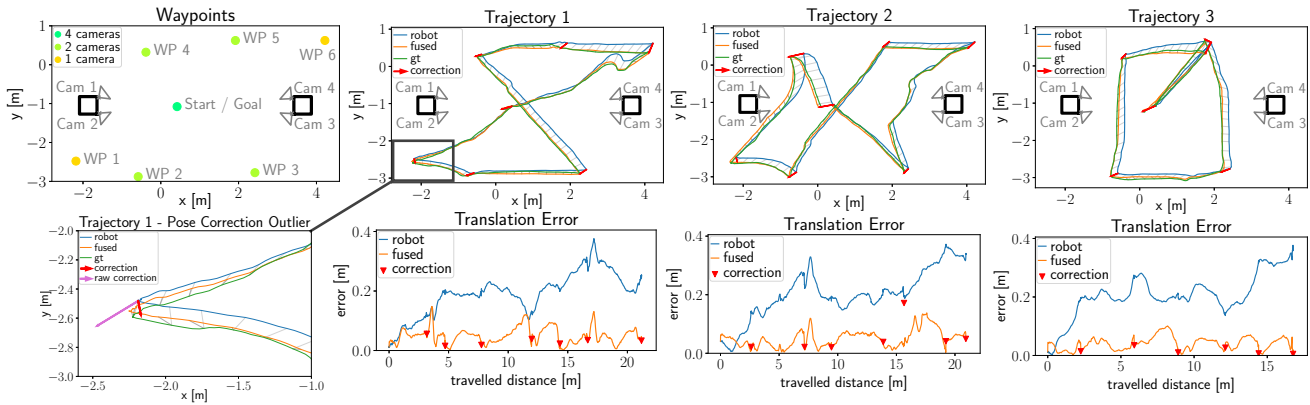

	\vspace{-1.5em}
	\centering
	\begin{tikzpicture}
		\node[inner sep=0,anchor=north west] (image1) at (0, 0){\includegraphics[height=3.7cm]{figures/2022-09-02_poses_7_traj2r_waypoints.pdf}};
		\node[inner sep=0,anchor=north west,xshift=-0.4cm, yshift=-0.12cm] (image2) at (image1.north east) {\includegraphics[height=3.5cm]{figures/2022-09-02_poses_11_traj1_trajectory.pdf}};
		\node[inner sep=0,anchor=north west,xshift=-0.4cm, yshift=-0.0cm] (image3) at (image2.north east) {\includegraphics[height=3.5cm]{figures/2022-09-02_poses_7_traj2r_trajectory.pdf}};
		\node[inner sep=0,anchor=north west,xshift=-0.4cm,yshift=-0.03cm] (image4) at (image3.north east) {\includegraphics[height=3.5cm]{figures/2022-09-02_poses_13_traj3r_trajectory.pdf}};
		
		\draw[line width=0.35mm, darkgray] (5.16, -2.9) rectangle ++(.8,.53);
		\draw[line width=0.35mm, darkgray] (5.16, -2.9) -- (4.05, -3.58);
		
		\node[inner sep=0,anchor=north,yshift=.7cm] (image5) at (image1.south) {\includegraphics[height=3.cm]{figures/2022-09-02_poses_11_traj1_trajectory_outlier.pdf}};
		\node[inner sep=0,anchor=north,xshift=-0.15cm, yshift=1.1cm] (image6) at (image2.south) {\includegraphics[height=3.6cm]{figures/2022-09-02_poses_11_traj1_transl_error.pdf}};
		\node[inner sep=0,anchor=north,xshift=-0.07cm,yshift=1.cm] (image7) at (image3.south) {\includegraphics[height=3.5cm]{figures/2022-09-02_poses_7_traj2r_transl_error.pdf}};
		\node[inner sep=0,anchor=north,yshift=1.cm] (image8) at (image4.south) {\includegraphics[height=3.5cm]{figures/2022-09-02_poses_13_traj3r_transl_error.pdf}};
	\end{tikzpicture}
	\vspace{-3.em}
	\caption{Evaluation of robot pose estimation: top-left: Room-setup with camera positions and waypoints colored by number of cameras from which waypoints are visible; center and right: plots of one resp. iteration of the three trajectories used for evaluation of the translation error over the traveled distance; bottom-left: example for outlier detection at WP~1 observed from only a single camera. The pose correction from the external cameras is not sent as feedback to the robot during these experiments to measure the deviation over the full trajectory length. See \refsec{sec:pose_acc} for further explanations.}
	\label{fig:traj_eval}
\end{figure*}
\begin{table*}[t]
\vspace{-.5em}
\caption{Translation and Orientation error (mean $\pm$ std) at waypoints, by no. of cameras with robot dets. and pose estimation source.}
\vspace{-1.em}
\label{tab:pose_error}
\centering
\renewcommand{\arraystretch}{1.1} %
\linespread{0.95}\selectfont
\setlength{\tabcolsep}{3.3pt}
\sisetup{separate-uncertainty}
\begin{threeparttable}
\begin{tabular}{L{2.3cm}|cc|cc|cc|cc}
  \toprule
   Pose Estimation & \multicolumn{2}{c|}{1 Camera} & \multicolumn{2}{c|}{2 Cameras} & \multicolumn{2}{c|}{4 Cameras} & \multicolumn{2}{c}{\textbf{Average}}\\
  \midrule
   Robot           &
   \SI[multi-part-units = single]{20.6\pm 7.6}{\centi\meter} &
   \SI[multi-part-units = single]{1.17\pm 1.19}{\degree} &
   \SI[multi-part-units = single]{17.1\pm 6.9}{\centi\meter} &
   \SI[multi-part-units = single]{1.03\pm 1.23}{\degree} &
   \SI[multi-part-units = single]{25.0\pm 6.9}{\centi\meter} &
   \SI[multi-part-units = single]{1.30\pm 1.96}{\degree} &
   \SI[multi-part-units = single]{19.1\pm 7.6}{\centi\meter} &
   \SI[multi-part-units = single]{1.11\pm 1.38}{\degree}\\
   
   Cameras (raw)   &
   \SI[multi-part-units = single]{13.8\pm 10.1}{\centi\meter} &
   \SI[multi-part-units = single]{3.39\pm 2.87}{\degree} &
   \SI[multi-part-units = single]{2.59\pm 1.49}{\centi\meter} &
   \SI[multi-part-units = single]{0.97\pm 0.79}{\degree} &
   \SI[multi-part-units = single]{2.88\pm 1.42}{\centi\meter} &
   \SI[multi-part-units = single]{1.02\pm 0.76}{\degree} &
   \SI[multi-part-units = single]{4.65\pm 6.22}{\centi\meter} &
   \SI[multi-part-units = single]{1.44\pm 1.72}{\degree}\\
   
   Cameras (1 frame) &
   \SI[multi-part-units = single]{8.23\pm 5.23}{\centi\meter} &
   \SI[multi-part-units = single]{1.40\pm 1.31}{\degree} &
   \SI[multi-part-units = single]{2.59\pm 1.49}{\centi\meter} &
   \SI[multi-part-units = single]{0.97\pm 0.79}{\degree} &
   \SI[multi-part-units = single]{2.88\pm 1.42}{\centi\meter} &
   \SI[multi-part-units = single]{1.02\pm 0.76}{\degree} &
   \SI[multi-part-units = single]{3.65\pm 3.36}{\centi\meter} &
   \SI[multi-part-units = single]{1.06\pm 0.92}{\degree}\\
   
   Cameras (5 frames) &
   \SI[multi-part-units = single]{7.77\pm 5.47}{\centi\meter} &
   \SI[multi-part-units = single]{1.18\pm 1.27}{\degree} &
   \textbf{\SI[multi-part-units = single]{2.58\pm 1.48}{\centi\meter}} &
   \SI[multi-part-units = single]{0.86\pm 0.68}{\degree} &
   \SI[multi-part-units = single]{2.82\pm 1.43}{\centi\meter} &
   \textbf{\SI[multi-part-units = single]{0.97\pm 0.74}{\degree}} &
   \SI[multi-part-units = single]{3.54\pm 3.31}{\centi\meter} &
   \SI[multi-part-units = single]{0.94\pm 0.84}{\degree} \\
   
   Fused          &
   \textbf{\SI[multi-part-units = single]{4.25\pm 1.57}{\centi\meter}} &
   \textbf{\SI[multi-part-units = single]{1.12\pm 1.08}{\degree}} &
   \SI[multi-part-units = single]{2.64\pm 1.48}{\centi\meter} &
   \textbf{\SI[multi-part-units = single]{0.79\pm 0.65}{\degree}} &
   \textbf{\SI[multi-part-units = single]{2.73\pm 1.41}{\centi\meter}} &
   \textbf{\SI[multi-part-units = single]{0.97\pm 0.76}{\degree}} &
   \textbf{\SI[multi-part-units = single]{2.93\pm 1.60}{\centi\meter}} &
   \textbf{\SI[multi-part-units = single]{0.88\pm 0.77}{\degree}}\\
   
   \bottomrule
\end{tabular}
\end{threeparttable}
\vspace{-1.5em}
\end{table*}
\subsection{Experiment Setup}
During the experiments, the HSR robot operates in a challenging, cluttered indoor environment of $\sim$\SI{240}{\square\meter} size.
Four external smart edge sensors are mounted at $\sim$\SI{2.5}{\meter} height in the center of the room to initialize and correct the robot localization.
As a reference for pose estimation, we employ the affordable and easy-to-use HTC Vive Pro tracking system, which was shown to yield position accuracies within a few millimeters~\cite{vive_tracker_2021}. For this, we place an HTC tracker on the robot's head, as shown in \reffig{fig:experimental_setup}.
For evaluation of the pose estimation accuracy, we define seven waypoints in the area observed by the external cameras and connect them in different ways to three different trajectories (\cf\reffig{fig:traj_eval}). 
As the tracking system and the camera network don't operate in the same reference frame, trajectories are rigidly aligned via Procrustes analysis~\cite{sturm_iros2012_benchmark} before evaluation. 

\subsection{Pose Estimation Accuracy}
\label{sec:pose_acc}
During a first set of experiments, we record a dataset of three iterations of each trajectory in both forward and reverse order, resulting in 18 sample trajectories. 
At the beginning of each trajectory, the robot's pose is initialized from observations of all four cameras and we verify that robot observations are consistent with the global model (\cf\reffig{fig:teaser}~(c)) to confirm a good initialization.
The 2D grid map used by the robot for LiDAR navigation is initialized with the prior model of the empty room.
The pose correction from external cameras is not sent to the robot during these experiments to measure the deviation over the full trajectory.

\reffig{fig:traj_eval} shows an exemplary iteration of each trajectory, comparing the path estimated by the robot's internal navigation stack and the fused trajectory estimate, obtained using pose corrections from the external cameras and pose graph optimization, with the ground-truth obtained from the HTC reference tracking system. Furthermore, the evolution of translation error is shown over the traveled distance.
While the robot's internal localization quickly accumulates errors of \SIrange{20}{30}{\centi\meter}, the error of the external camera pose estimation stays below \SI{5}{\centi\meter}, when the robot is observed from at least two cameras. The error can be higher when observing the robot in only a single camera, e.g. WP~1 (1st WP of Traj~1, resp. 5th WP of Traj.~2), but always improves upon the robot's internal estimate. The single-camera observations can further be improved by fusing with robot odometry via the pose graph.
In \reffig{fig:traj_eval}~(bottom-left), we illustrate the outlier detection for pose estimation from a single camera: The raw pose estimate shows an unrealistically high change in distance to the observing camera; therefore, the pose correction is restricted to the lateral direction.
\begin{table}[b]
\vspace{-1.5em}
\caption{Root mean square translation error in cm.}
\vspace{-1em}
\label{tab:ate_dataset}
\centering
\renewcommand{\arraystretch}{1.} %
\linespread{0.95}\selectfont
\setlength{\tabcolsep}{3pt}
\begin{threeparttable}
\begin{tabular}{L{2cm}|ccc|c}
  \toprule
  Pose Estimation & Traj. 1 & Traj. 2 & Traj. 3 & \textbf{Avg.}\\
  \midrule
  Robot & 20.5 & 19.4 & 15.8 & 18.5\\
  Fused & \textbf{5.76} & \textbf{4.64} & \textbf{3.03} & \textbf{4.48}\\
  \bottomrule
\end{tabular}
\end{threeparttable}
\end{table}

\reftab{tab:pose_error} reports a quantitative evaluation of the translation and orientation errors at the waypoints, ordered by the number of cameras from which the robot is observed.
The pose error is lowest when the robot is observed from two or four cameras and amounts to \SI{2.93}{\centi\meter} and \SI{0.88}{\degree} averaged over all waypoints, significantly improving over the robot internal localization with an average error of \SI{19.1}{\centi\meter} and \SI{1.11}{\degree}. The outlier detection for single-camera pose estimation significantly improves the accuracy in the resp. waypoints by \SI{5.6}{\centi\meter} resp. \SI{2.}{\degree} w.r.t. the raw estimate and prevents a worsening of the orientation estimate w.r.t. the robot's internal localization.
Averaging the pose estimates of multiple frame-sets and fusion with the robot odometry via the pose graph give further improvements.

\reftab{tab:ate_dataset} shows the root mean square (RMS) of the translation error for the trajectories. The shorter Traj.~3, without WP~1 and WP~6 visible in only one camera, has the lowest trajectory error. The fused trajectory estimate using external camera pose estimation and robot odometry gives a significant improvement from \SI{18.5}{\centi\meter} to \SI{4.48}{\centi\meter}, averaged over the dataset.

\begin{figure}[t]
	\vspace{-4.6em}
	\centering
	\includegraphics[width=0.9\linewidth]{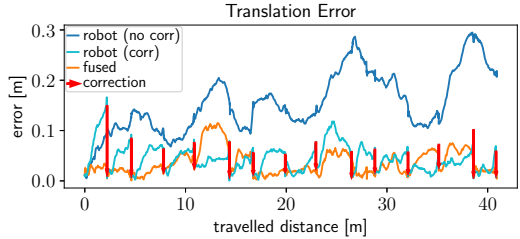}
	\vspace{-4.1em}
	\caption{Translation error with and without applying pose correction feedback to the robot's localization for the second experiment.}
	\label{fig:traj_fb}
	\vspace{-1.6em}
\end{figure}

In a second set of experiments, we compare the fused trajectory calculated on the central backend with the robot's internal estimate when integrating the pose correction feedback at static waypoints. For this, we record two iterations of a longer trajectory ($\sim$\SI{40}{\meter}, 3-times Traj.~3), with and without applying the feedback on the robot.
\reffig{fig:traj_fb} shows the translation error over the traveled distance. %
The pose correction feedback significantly improves the robot's localization, reducing the error to the order of magnitude of the fused path calculated on the backend. 
Comparing the RMS translation error of the complete trajectories, the robot estimate without correction amounts to \SI{17.5}{\centi\meter} and is improved significantly to \SI{5.51}{\centi\meter} by the pose correction feedback.
The pose graph fusion further improves the RMSE to \SI{3.34}{\centi\meter}.
To integrate the pose graph fusion on the robot, a closer coupling with the backend or a re-implementation of the robot navigation stack would be required.

\subsection{Collaborative Semantic Mapping}
\begin{figure}[b]
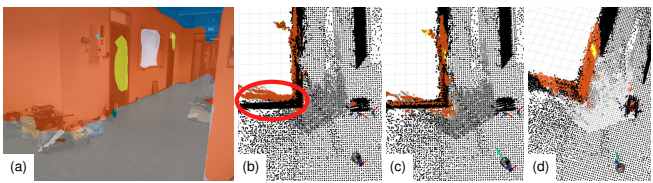

\vspace{-1.em}
	\centering
	\begin{tikzpicture}
		\node[inner sep=0,anchor=north east, xshift=-0.05cm] (image0) at (image1.north west) {\includegraphics[height=2.3cm,trim=0 0 0 0,clip]{figures/map_coherency_2D_view.png}};
		\node[inner sep=0,anchor=north west] (image1) at (0, 0) {\includegraphics[height=2.3cm,trim=120 30 60 20,clip]{figures/map_coherency_bad_2.png}};
		\draw[red!90, ultra thick] (0.45,-1.25) ellipse (.47 and .25);
		\node[inner sep=0,anchor=north west,xshift=0.05cm] (image2) at (image1.north east) {\includegraphics[height=2.3cm,trim=130 30 50 20,clip]{figures/map_coherency_good.png}};
		\node[inner sep=0,anchor=north west,xshift=0.05cm] (image3) at (image2.north east) {\includegraphics[height=2.3cm,trim=30 30 70 20,clip]{figures/map_coherency_fused.png}};
		
		\node[label,scale=.75, anchor=south west, rectangle, fill=white, align=center, font=\scriptsize\sffamily] (n_0) at (image0.south west) {(a)};
			\node[label,scale=.75, anchor=south west,xshift=-0.05cm, rectangle, fill=white, align=center, font=\scriptsize\sffamily] (n_1) at (image1.south west) {(b)};
			\node[label,scale=.75, anchor=south west,xshift=-0.05cm, rectangle, fill=white, align=center, font=\scriptsize\sffamily] (n_3) at (image2.south west) {(c)};
			\node[label,scale=.75, anchor=south west,xshift=-0.05cm, rectangle, fill=white, align=center, font=\scriptsize\sffamily] (n_3) at (image3.south west) {(d)};
	\end{tikzpicture}
	\vspace{-1.7em}
	\caption{Consistency of robot observations with global map: (a) local robot view; (b) 3D view with accumulated drift after Traj. 3 and (c) after pose correction (green arrow); (d) robot observations fused into map.}
	\label{fig:map_coherency}
\end{figure}

In further experiments, we demonstrate the integration of the robot as a mobile sensor node into the smart edge sensor network.
\reffig{fig:map_coherency} shows the consistency of the robot observations with the allocentric scene model.
Without pose correction feedback, error accumulates in the localization and the observations have low consistency with the model. After pose correction through the feedback from the smart edge sensors, the observations fit the model well and can consistently be fused into the semantic map.

\begin{figure}[t]
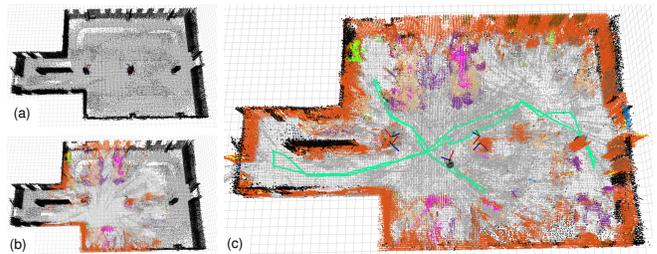

	\centering
	\begin{tikzpicture}
		\node[inner sep=0,anchor=north west] (image1) at (0, 0) {\includegraphics[height=1.63cm,trim=30 100 50 100,clip]{figures/map_init.png}};
		\node[inner sep=0,anchor=north west,yshift=-0.1cm] (image2) at (image1.south west) {\includegraphics[height=1.63cm,trim=20 20 0 45,clip]{figures/map_static_sensors.png}};
		\node[inner sep=0,anchor=north west,xshift=0.1cm] (image3) at (image1.north east) {\includegraphics[height=3.36cm,trim=50 50 30 50,clip]{figures/map_explored.png}};
		
		\node[label,scale=.75, anchor=south west, rectangle, fill=white, align=center, font=\scriptsize\sffamily] (n_0) at (image1.south west) {(a)};
			\node[label,scale=.75, anchor=south west,xshift=-0.07cm, rectangle, fill=white, align=center, font=\scriptsize\sffamily] (n_1) at (image2.south west) {(b)};
			\node[label,scale=.75, anchor=south west,xshift=-0.07cm, rectangle, fill=white, align=center, font=\scriptsize\sffamily] (n_3) at (image3.south west) {(c)};
	\end{tikzpicture}
	\vspace{-2.em}
	\caption{Completion of semantic map by fusing with robot observations: (a) initial 3D map, (b) incomplete semantic map with observations from static smart edge sensors, (c) semantic map completed with robot observations and exploration path.}
	\label{fig:map_exploration}
	\vspace{-.5em}
\end{figure}

We show the collaborative semantic mapping in a lab-scale experiment in \reffig{fig:map_exploration}. The semantic map is initialized from a prior model of the empty room, without any semantic information. The static smart edge sensors observe the room only partly, due to occlusions and limited measurement range, providing semantic information for $\sim$\SI{50}{\percent} of the voxels.
The mobile robot provides changing sensor perspectives and can actively perceive the areas not observed by the static sensors. The waypoints for completing the map were defined manually in the experiment but could be set automatically using approaches for exploration and viewpoint optimization~\cite{dong2019multi}.
Through collaboration, mobile robot and static sensors build a complete semantic map of the $\sim$\SI{240}{\square\meter} environment with semantic information for over \SI{90}{\percent} of the voxels.

\subsection{Localization Robustness}
To evaluate the robustness of our collaborative localization approach,
we repeat the lab-scale experiment ten times, five times with applying the pose correction feedback and five times without, and report the success rate in \reftab{tab:robustness}.
Using only the internal LiDAR-based localization, the robot cannot reach all waypoints in three of five trials, an emergency stop due to collision with an obstacle occurred in one trial, and, hence, the success rate reaches only 1 / 5. Failures occur after \SI{38}{\m} of traveled distance, on average.

With the proposed localization feedback, the robot completes the $\sim$\SI{60}{\meter} long trajectory successfully in all trials.
The external camera-based pose estimation compensates for the difficulties of the robot-internal navigation to localize in the highly cluttered, dynamic environment where only few distinctive features, such as walls or columns, are visible in the LiDAR, significantly increasing the system's robustness.
\begin{table}[t]
\vspace{-.6em}
\caption{Robustness during 5+5 lab-scale experiments.}
\vspace{-1.em}
\label{tab:robustness}
\centering
\renewcommand{\arraystretch}{1.} %
\linespread{0.95}\selectfont
\setlength{\tabcolsep}{3pt}
\begin{threeparttable}
\begin{tabular}{L{2.2cm}|cc|c}
  \toprule
   & WP not reached & Emergency Stop & Success Rate\\
  \midrule
  w/o correction fb & 3 / 5 & 1 / 5 & 1 / 5\\
  w/ correction fb & \textbf{0 / 5} & \textbf{0 / 5} & \textbf{5 / 5}\\
  \bottomrule
\end{tabular}
\end{threeparttable}
\vspace{-1.65em}
\end{table}
 
\section{Conclusions}
\label{sec:Conclusion}
We presented a novel method for marker-less mobile robot pose estimation using multi-view keypoint detections from a network of external smart cameras. 
We use this to initialize and continuously update a mobile robot's localization in the allocentric scene model of the smart edge sensor network and build a system for collaborative perception between mobile robot and static smart sensors.
The typical position error of our proposed method for mobile robot pose estimation is below \SI{2.8}{\centi\meter} when detected in at least two cameras, and below \SI{4.3}{\centi\meter} when detected in a single camera, while the robot localization typically deviates more than \SI{19}{\centi\meter} after only \SI{5}{\meter} of traveled distance.

Precisely initializing and tracking the robot's localization w.r.t the camera network allows to fuse its semantic observations in a globally consistent way into the allocentric scene model.
The robot as a mobile sensor node provides changing viewpoints and can explore areas not covered by the static sensors due to occlusions and limited measurement range.
We demonstrate a real-world application where a mobile robot and distributed smart edge sensors collaboratively build a 3D semantic map of a large room. %

In future work, we plan to use our system for collaborative perception to enable anticipative human-aware navigation and human-robot interaction in a shared workspace.

\section*{Acknowledgement}
The authors would like to thank Jonas Bode and Julian Hau for their help with preparing the training data.

\IEEEtriggeratref{21}
\bibliographystyle{IEEEtran}
\bibliography{literature}

\end{document}